\documentclass[conference]{IEEEtran}
\IEEEoverridecommandlockouts

\usepackage{cite}
\usepackage{amsmath,amssymb,amsfonts}
\usepackage{algorithmic}
\usepackage{graphicx}
\usepackage{textcomp}
\usepackage{colortbl}
\usepackage{subfig} 
\usepackage{float} 
\usepackage[table]{xcolor}
\usepackage{xcolor}
\begin{document}

\def\x{{\mathbf x}}
\def\L{{\cal L}}
\newcommand{\red}{\color{red}}

\makeatother

\title{Learning Unbiased Image Segmentation: A Case Study with Plain Knee Radiographs}

\makeatletter
\newcommand{\linebreakand}{%
  \end{@IEEEauthorhalign}
  \hfill\mbox{}\par
  \mbox{}\hfill\begin{@IEEEauthorhalign}
}


%
\makeatother
\author{\IEEEauthorblockN{Nickolas Littlefield} 
\IEEEauthorblockA{\text{\textit{University of Pittsburgh}}\\\textit{ngl18@pitt.edu}}
\and\IEEEauthorblockN{Johannes F. Plate, MD, PhD} 
\IEEEauthorblockA{
\text{\textit{University of Pittsburgh}}\\\textit{johannes.plate@pitt.edu}}
\and\IEEEauthorblockN{Kurt R. Weiss, MD} 
\IEEEauthorblockA{\text{\textit{University of Pittsburgh}}\\\textit{krw13@pitt.edu}}
\and \IEEEauthorblockN{Ines Lohse, PhD} 
\IEEEauthorblockA{
\text{\textit{\textit{University of Pittsburgh}}}\\\textit{inl22@pitt.edu}}
\linebreakand
\and\IEEEauthorblockN{Avani Chhabra} 
\IEEEauthorblockA{\text{\textit{University of Pittsburgh}}\\\textit{avc41@pitt.edu}}
\and\IEEEauthorblockN{Ismaeel A. Siddiqui} 
\IEEEauthorblockA{
\text{\textit{University of Pittsburgh}}\\\textit{iss80@pitt.edu}}
\and\IEEEauthorblockN{Zoe Menezes} 
\IEEEauthorblockA{\text{\textit{University of Pittsburgh}}\\\textit{zim9@pitt.edu}}
\and \IEEEauthorblockN{George Mastorakos, MD, MS} 
\IEEEauthorblockA{
\text{\textit{Cortechs.ai}}\\\textit{georgemasto@gmail.com}}
\linebreakand

\and\IEEEauthorblockN{Sakshi Mehul Thakar} 
\IEEEauthorblockA{\text{\textit{University of Pittsburgh}}\\\textit{smt150@pitt.edu}}
\and\IEEEauthorblockN{Mehrnaz Abedian} 
\IEEEauthorblockA{
\text{\textit{University of Pittsburgh}}\\\textit{mea209@pitt.edu}}
\and\IEEEauthorblockN{Matthew F. Gong, MD} 
\IEEEauthorblockA{\text{\textit{University of Pittsburgh}}\\\textit{gongm2@upmc.edu}}
\and \IEEEauthorblockN{Luke A. Carlson, MS} 
\IEEEauthorblockA{
\text{\textit{University of Pittsburgh}}\\\textit{lac249@pitt.edu}}
\linebreakand
\and\IEEEauthorblockN{Hamidreza Moradi, PhD*} 
\IEEEauthorblockA{\text{\textit{University of Mississippi Medical Center}}\\\textit{hmoradi@umc.edu}} 
\and\IEEEauthorblockN{Soheyla Amirian, PhD*} 
\IEEEauthorblockA{\text{\textit{University of Georgia}}\\\textit{amirian@uga.edu}}

\and\IEEEauthorblockN{Ahmad P. Tafti, PhD*\thanks{*Corresponding authors, and equal contributions.}} 
\IEEEauthorblockA{\text{\textit{University of Pittsburgh}}\\\textit{tafti.ahmad@pitt.edu}}
}
\maketitle

\begin{abstract}
Automatic segmentation of knee bony anatomy is essential in orthopedics, and it has been around for several years in both pre-operative and post-operative settings. While deep learning algorithms have demonstrated exceptional performance in medical image analysis, the assessment of fairness and potential biases within these models remains limited. This study aims to revisit deep learning-powered knee-bony anatomy segmentation using plain radiographs to uncover visible gender and racial biases. The current contribution offers the potential to advance our understanding of biases, and it provides practical insights for researchers and practitioners in medical imaging. The proposed mitigation strategies mitigate gender and racial biases, ensuring fair and unbiased segmentation results. Furthermore, this work promotes equal access to accurate diagnoses and treatment outcomes for diverse patient populations, fostering equitable and inclusive healthcare provision.
\end{abstract}

\begin{IEEEkeywords}
Image Segmentation, AI Fairness, Knee Radiographs, Unbiased Image Segmentation, Safe AI
\end{IEEEkeywords}

\section{Introduction}
\label{sec:intro}
Automatic segmentation of knee bony anatomy is vital in orthopedics for precise information about the knee joint's structure. It aids in pre-operative planning, post-operative evaluation, knee alignment analysis, and disease progression monitoring, to name a few. Recent years have witnessed remarkable progress in medical image analysis, particularly through the use of deep learning computer vision algorithms. These algorithms employ complex Artificial Neural Networks (ANNs) to extract valuable features from extensive knee image datasets, enabling accurate identification and segmentation of various bony structures within the knee joint \cite{10.1007/978-3-030-64559-5_12,goswami2023automatic,sagheb2021leveraging,yan2022knee,amirian2022word,kulseng2023automatic}.

Despite their impressive capabilities, there is a growing concern regarding the fairness and potential biases within these deep learning models \cite{toutiaee2020stereotypefree,lokhande2020fairalm,du2020fairness,tripathi2023fairness}. Fairness in the context of medical image analysis refers to the equitable performance of the deep learning/AI algorithm across different patient populations. Biases, on the other hand, refer to systematic errors or inconsistencies in the algorithm's predictions that may disproportionately affect certain groups of patients. It is now essential to assess fairness and potential biases in deep learning models used for knee bone segmentation. This evaluation is crucial for promoting equitable healthcare delivery and preventing disparities in patient outcomes, thereby enhancing the provision of equitable and inclusive healthcare in the field of orthopedics. However, the assessment of fairness and biases in these models is currently limited. It is thus crucial to thoroughly investigate these issues and develop computational methods that can analyze and subsequently mitigate biases. These measures are necessary to ensure consistent and accurate performance of the algorithms across diverse patient populations.

To address the assessment of fairness, researchers and practitioners need to consider factors such as the representation of diverse populations within the training data, potential sources of bias in the data collection process, and the impact of demographic factors on the algorithm's performance \cite{sikstrom2022conceptualising,chen2021algorithm}. By carefully analyzing these aspects, it becomes possible to identify and rectify biases that may arise from the data or algorithm design.
Moreover, the transparency and interpretability of deep learning models also play a significant role in addressing fairness and biases. Understanding how the algorithm arrives at its predictions can help uncover any biases or limitations in the model's decision-making process. Techniques such as explainable AI and model interpretability methods can aid in identifying potential sources of bias and ensuring that the algorithm's predictions are fair and unbiased.

While deep learning algorithms have demonstrated exceptional performance in knee-bony anatomy segmentation and other medical image analysis tasks, it is crucial to evaluate their fairness and potential biases. By tackling these issues, we aim to ensure that these algorithms deliver accurate and unbiased results, thereby enhancing patient care and improving outcomes in orthopedics. Our study makes significant research contributions by:

\begin{itemize}
    \item Revisiting the utilization of deep learning for knee bony anatomy segmentation using plain radiographs, specifically to uncover visible biases related to gender and race.

    \item Proposing effective strategies to mitigate gender and racial biases, ensuring that the segmentation results are fair and unbiased.

    \item Providing valuable practical insights that can benefit researchers and practitioners in the field of medical imaging.

    \item Promoting equitable and inclusive healthcare provision by enabling equal access to accurate diagnoses and treatment outcomes for diverse patient populations.
    
\end{itemize}
The organization of the current contribution is as follows. 
Section \ref{sec:MM} presents the materials and methods employed in this study. Experimental validation and model evaluation are demonstrated in Section \ref{sec:EV}. Section \ref{sec:Conclusion} first discusses the findings and implications of the work and subsequently concludes the study while offering insights for future research.

\section{Materials and Methods}
\label{sec:MM}
This section aims to present a comprehensive overview of our study design. Initially, we will present the dataset utilized for segmenting knee radiographs. Subsequently, the segmentation model employed in this study will be briefly discussed, followed by delving into the details of the techniques examined to analyze and mitigate bias in the modeling. Finally, we will discuss the fairness evaluation metrics.

\subsection{Dataset Procurement and Preprocessing}

This study utilizes the Osteoarthritis Initiative (OAI) data repository, a well-established research and educational resource supported by the National Institute of Health (NIH) \cite{oaidataset}. The dataset is employed to randomly select 403 radiographs/patients for the purpose of manually annotating different knee segments. 
The specific elements of knee segments included in this study are the Femur, Tibia, Fibula, and Patella. 
To assess inter-rater agreement for manual segmentation, three image annotators were provided with 20 identical images in two separate batches (10 images in each batch). The agreement among annotators was satisfactory, with an average Intersection over Union (IoU) of 0.873 and an average Dice coefficient of 0.9145. Manual segmentation was performed using ITK-SNAP \cite{yushkevich2016itk}. The annotated data was then divided into train, validation, and test sets, comprising 70\%, 15\%, and 15\% of the data, respectively. By utilizing the annotated data, the deep learning models developed in this study aim to achieve robust and accurate analysis of knee joint segmentation in radiographs. The models learn and identify critical patterns necessary for precise segmentation, resulting in enhanced performance. 
Figure~\ref{fig:segmentation} provides the radiographs for two patients selected randomly, with their corresponding manual segmentation masks, along with the masks predicted by the algorithm.

\begin{figure}[!ht]
    \begin{center}
    \includegraphics[width=1.0\linewidth]{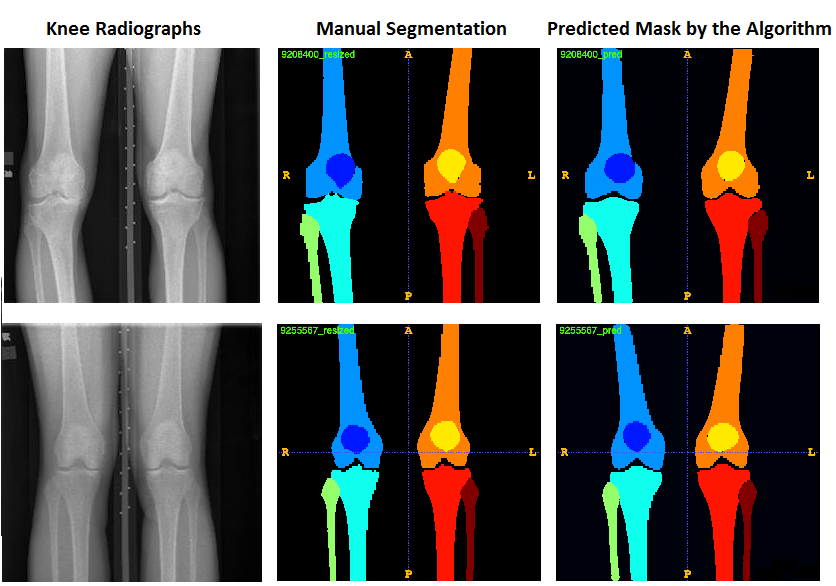}
    \caption{Exemplar Patient Radiographs (Left), the manual segmentation masks (middle), and the predicted segmentation masks by the algorithm (right). We achieved IoU of 0.762 and 0.781 for the first and second rows, respectively.} 
    \vspace{-0.05in}
    \label{fig:segmentation}
    \end{center}
\end{figure}

\subsection{Deep Learning for Segmentation}

Image segmentation is an essential task in the field of computer vision and finds widespread application in the analysis of medical images. In recent years, deep learning models have emerged as state-of-the-art methods for achieving highly accurate image segmentation. Among these models, U-Net~\cite{unet} has gained popularity in the medical domain due to its ability to effectively capture spatial information and learn complex image features. The U-Net architecture, based on an encoder-decoder design, incorporates skip connections to preserve high-resolution features while reducing the number of trainable parameters.

In this study, we trained a U-Net model with a ResNet18~\cite{resnet} backbone for the segmentation of knee bones and patella. The training and optimization of the model were conducted using Python 3.10.11 and PyTorch 1.13.1. The model was trained using the available training and validation datasets and subsequently evaluated on an independent test set. 
To assess the accuracy of the model and evaluate its fairness, the IoU scores for each segmented region were employed as performance metrics.

 \subsection{Baseline Model and Bias Mitigation Techniques}

 In order to evaluate the efficacy of various methods for analyzing and mitigating bias, we employed a baseline model and explored several techniques as detailed below. Subsequent sections will present an assessment of their influence on the fairness of the model.

\subsubsection{Baseline Model} Recognizing the potential presence of bias in deep learning/machine learning models, we aimed to establish a baseline for comparison prior to delving into diverse bias mitigation techniques. To achieve this, we initially trained the model using the complete training dataset, without applying any bias mitigation techniques. This approach ensures that the distribution of protected attributes, such as gender and race, in the training set accurately represents the distribution in the entire dataset. For the purposes of this study, we will refer to this model as the  \textit{baseline model}.

\subsubsection{Dataset Balancing} To address potential bias and enhance the model's ability to accurately classify examples from underrepresented groups, we employed a dataset balancing technique. Specifically, we performed oversampling of the underrepresented protected groups in the training data. This approach aimed to mitigate the risk of the model exhibiting bias towards the majority group and improve its capacity to effectively classify instances from underrepresented groups. We will refer to this model as the  \textit{Balanced model}.

\subsubsection{Stratified Batching} To ensure a balanced representation of the protected attributes of interest during model training in each batch, we utilized a technique called stratified batching. This approach involved grouping examples based on these attributes and randomly sampling instances from each group to create mini-batches for training. By employing this method, we ensured that the model was trained on an unbiased distribution of examples from each protected group within every training batch, promoting fairness in the learning process. We will refer to this model as the  \textit{Stratified model}.

\subsubsection{Group-Specific Models} As an additional approach to mitigate bias, we investigated the concept of training individual models for each protected group. By creating distinct models tailored to each group, this strategy has the potential to enhance performance specifically for underrepresented groups. By exploring this avenue, we aimed to examine the effectiveness of group-specific models in mitigating bias and improving overall model fairness, referred to as the \textit{Group-Specific Model}

The effectiveness of these techniques was assessed by measuring the models' performance on the entire dataset and comparing the results with those of the baseline model. This evaluation process allowed us to gauge the impact of the applied techniques and determine their efficacy in promoting fairness and reducing bias within the models.

\subsection{Quantifying Bias and Ensuring Fairness} 
For the evaluation of our models, we utilize two well-established statistical measures, namely the Standard Deviation (SD) and the Skewed Error Ratio (SER). These measures play a crucial role in gaining insights into the distribution of average IoU values across different groups, helping us identify variations and imbalances.

The SD serves as a metric for quantifying the extent of data dispersion from its mean value. By assessing the SD of errors, we can gain an understanding of how IoU values are scattered and deviated from the average, allowing us to explore the error variation within and among protected groups.

On the other hand, SER focuses specifically on evaluating the potential bias in the algorithm's prediction errors toward specific groups. By calculating the ratio of the highest to the lowest error rate among protected groups. This measure enables us to detect any disparities or skewed tendencies in the algorithm's predictions, uncovering biases. 



\section{Experimental Validation}
\label{sec:EV}

To comprehensively assess the fairness of our segmentation model across various protected subgroups, we undertook a series of experiments. Our approach involved sampling the training data while taking into account a protected attribute, resulting in diverse training dataset configurations.
The significance of these experiments lies in their ability to validate the fairness of our segmentation models.
Specifically, we focused on two protected attributes, namely \textit{race} and \textit{gender}, and separately examined the effectiveness of mitigation strategies for each.

\subsection{Experimental Setup}

For training, validating, and testing our models, we leveraged Jupyter Notebook sessions using the Data Science service on Oracle Cloud that was equipped with an Nvidia V100 GPU with 16GB of memory. The implementation of our models was carried out using Python 3.10.11 and PyTorch 1.13.1.
To achieve high segmentation accuracy, we conducted numerous experiments to identify the optimal hyperparameters. Eventually, our models were trained for 50 epochs, employing a batch size of 16. We employed the Adam optimizer with a learning rate of 0.0004. Throughout all the segmentation experiments, we employed an IoU threshold of 0.5 as a criterion to gauge the models' accuracy.

\subsection{Qualitative and Quantitative Evaluation}


Table~\ref{tab:distribution} presents the distribution of data across different gender and racial groups within the study as protected attributes. The table reveals that the sample predominantly consisted of female participants, accounting for 60\% of the total. Additionally, it highlights that the female group had a higher representation of Black or African American individuals compared to the male group, with 57\% of female participants being identified as African American, while only 43\% of male participants belonged to the same racial group. 
Nevertheless, it is evident that the race distribution, when considered as a whole, exhibits a striking similarity, with 48\% being white and the remaining portion composed of black or African American individuals. These distributional insights are crucial for comprehending the fairness metrics and their variations depicted in Tables~\ref{tab:ethnical} and \ref{tab:gender}. 


In Tables~\ref{tab:ethnical}, we observe that the baseline model achieves high IoU scores for both White/Caucasian and Black/African American individuals, with values of 0.833 and 0.834, respectively. However, when applying the balanced model, intended to address biases, we see a slight decrease in performance for both racial groups, with IoU scores of 0.836 for White/Caucasian individuals and 0.832 for Black/African American individuals. This decrease suggests that simply balancing the dataset does not necessarily lead to improved fairness. In contrast, stratified modeling proves to be an effective technique for mitigating racial biases, as it yields lower SER values for both racial groups compared to the baseline model. 

Similarly, when considering gender groups (Table~\ref{tab:gender}) the baseline model exhibits high IoU scores of 0.836 for males and 0.813 for females. However, employing the balanced model results in lower IoU scores, increase prediction error variability, and a decrease in fairness. Interestingly, stratified modeling demonstrates better fairness outcomes for gender groups compared to the baseline model Although the accuracy is slightly compromised, the SER values indicate improved fairness. On the other hand, group-specific modeling seems to introduce larger disparities, indicating that this approach is less effective in mitigating gender biases.

These results highlight the trade-offs between fairness and accuracy when implementing different bias mitigation techniques. While stratified modeling shows promising fairness improvements, there is a slight compromise in accuracy. It emphasizes the importance of carefully selecting and evaluating the most suitable approach based on the specific context and priorities of the application. Further research is necessary to develop novel techniques that can achieve a better balance between fairness and accuracy, ensuring equitable outcomes for all individuals while maintaining high prediction performance.




\begin{table}[h!] 
\centering
\caption{Racial and Gender Groups Distribution.}
\label{tab:distribution}
\begin{tabular}{|c|c|c|c|}
\hline
\textbf{\cellcolor{blue!25}Sex} & \textbf{\cellcolor{blue!25}Race} & \textbf{\cellcolor{blue!25}Patients} &\textbf{\cellcolor{blue!25}Percentage} \\
\cline{1-4}
Male & White or Caucasian & 91 & 23\% \\
\cline{2-4}
& Black or African American & 71 & 17\% \\
\cline{1-4}
Female & White or Caucasian & 102 & 25\% \\
\cline{2-4}
& Black or African American & 139 & 35\% \\
\hline
\end{tabular}
\end{table}

\begin{table}[h!]
\centering
\caption{Racial Groups Average IoU \& Fairness Scores.}
\label{tab:ethnical}
\begin{tabular}{|l|c|c|c|c|}
\hline
\textbf{\cellcolor{blue!25}Model} & \textbf{\cellcolor{blue!25}White/ Caucasian} & \textbf{\cellcolor{blue!25}Black/ AA*}& \textbf{\cellcolor{blue!25}SD}& \textbf{\cellcolor{blue!25}SER} \\ \hline
Baseline & 0.833 & 0.834 & 0.001 & 1.004\\ \hline
Balanced & 0.836 & 0.832 & 0.002 & 1.023\\ \hline
Stratified & 0.768 & 0.767 & 0.001 & 1.003\\ \hline
Group-Specific & 0.797 & 0.801 & 0.002 & 1.020\\ \hline
\end{tabular}
\vspace{0.02in}
\caption*{\small *African American.}
\end{table}

\begin{table}[h!]
\centering
\caption{Gender Groups Average IoU \& Fairness Scores.}
\label{tab:gender}
\begin{tabular}{|l|c|c|c|c|}
\hline
\textbf{\cellcolor{blue!25}Model} & \textbf{\cellcolor{blue!25}Malel} & \textbf{\cellcolor{blue!25}Female}& \textbf{\cellcolor{blue!25}SD} & \textbf{\cellcolor{blue!25}SER} \\ \hline
Baseline & 0.836 & 0.813 & 0.015 & 1.137\\ \hline
Balanced & 0.804 & 0.765 & 0.027 & 1.196\\ \hline
Stratified & 0.714 & 0.716 & 0.001 & 1.006\\ \hline
Group-Specific & 0.742 & 0.793 & 0.036 & 1.250\\ \hline
\end{tabular}
\end{table}

\section{Discussion, Conclusion, and Outlook}
\label{sec:Conclusion}
   Our study revisited the implementation of deep learning in knee bony anatomy segmentation of plain radiographs to uncover gender and racial biases and implement strategies for bias mitigation. Within the multiple models we implemented, we found that different bias mitigation strategies present a compromise between fairness and accuracy of predicted knee anatomy segmentation. Optimizing a deep learning model that can fairly account for racial and gender bias in the interpretation of knee plain radiographs has significant implications in several areas. 

   As in many subspecialties, it is well-known in the field of orthopedics that bone diseases are not evenly distributed throughout the population. For example, osteosarcoma has a well-established male predominance (1.22:1) and higher incidence in African Americans (6.8 per million persons per year) and Hispanics (6.5 per million) compared to Caucasians (4.6 per million) \cite{ottaviani2010epidemiology}. Developing a model for knee plain radiograph interpretation with reduced bias can provide a means to identify and improve the quality of care for historically marginalized patient populations. Racial inequity in the performance of total knee arthroplasty has been demonstrated previously between Black/African American and White/Caucasian patients, despite Black patients having higher reported rates of clinical knee osteoarthritis \cite{wilson1994racial}. Factors involved in this discrepancy are complex, with Black patients found to be less willing to undergo surgery, exhibiting worse surgical outcomes, and undergoing surgery at lower-volume hospitals \cite{usiskin2022racial,thirukumaran2020geographic}. Surgeon bias against offering total knee arthroplasty likely also plays a significant role and could be further evaluated prospectively using an unbiased predictive model. Similar findings have been found in other groups including the Hispanic patient population \cite{escalante2000recipients}. Understanding the interplay of treatment offered for osteoarthritis with racial or ethnic minority status, socioeconomic status, and level of education requires further investigation. 
   In conclusion, our study found that racial and gender bias can be present in knee bony anatomy segmentation models, but that bias mitigation strategies can be effectively implemented. Application of models with reduced bias has utility in understanding how to improve inequitable treatment for clinical conditions including osteoarthritis. Our future work will focus on the uptake and clinical utility.

\section*{Acknowledgment}

The authors first declare that they have no competing interests. This work was supported in part by Oracle Cloud credits and related resources provided by Oracle for Research. Additionally, this work was also supported by the University of Pittsburgh Clinical and Translational Science Institute / National Institutes of Health (UL1TR001857). Any opinions, findings, and conclusions
or recommendations expressed in this material are those of
the author(s) and do not necessarily reflect the views of the Oracle for Research and/or the University of Pittsburgh Clinical and Translational Science Institute / National Institutes of Health. 

\bibliographystyle{IEEEbib}
\bibliography{refs}

\end{document}